\documentclass[11pt]{article}

\usepackage[showframe=false, margin=1.0in]{geometry}

\usepackage[utf8]{inputenc}
\usepackage{wrapfig}
\usepackage[lofdepth,lotdepth]{subfig}
\usepackage{booktabs}
\usepackage{array}
\usepackage{longtable}
\usepackage{rotating}
\usepackage{caption}
\usepackage{csquotes}
\usepackage{stfloats}
\usepackage[T1]{fontenc}
\usepackage{textcomp}
\usepackage{times}
\usepackage{tabularx}
\usepackage{soul}
\usepackage{csquotes}
\usepackage{url}

\usepackage{color}
\definecolor{black}{gray}{0} 

\title{
From Model Training to Model Raising\\
\vspace{2mm}
\normalsize{A call to reform AI model training paradigms\\ from post-hoc alignment to intrinsic, identity-based development}
}

\author{
Roland Aydin,$^{1,2}$
Christian Cyron,$^{1,2}$
Steve Bachelor,$^{3}$
Ashton Anderson,$^{4}$
Robert West$^{5}$\\
{\small $^{1}$Hamburg University of Technology, Germany}\\
{\small $^{2}$Helmholtz-Zentrum Hereon, Germany}\\
{\small $^{3}$Independent Researcher}\\
{\small $^{4}$University of Toronto, Canada}\\
{\small $^{5}$EPFL, Switzerland}
}

\date{22 June 2025}

\usepackage[utf8]{inputenc}
\usepackage[T1]{fontenc}
\usepackage{hyphenat}
\usepackage{xspace}
\usepackage{amsmath}
\usepackage{amsfonts}
\usepackage[colorlinks=true,linkcolor=black,citecolor=black]{hyperref}
\usepackage{booktabs}
\usepackage{multirow}
\usepackage{makecell}
\usepackage{caption}
\usepackage{minibox}
\usepackage{bbm}
\usepackage{graphicx}
\usepackage{balance}
\usepackage{mathtools}
\usepackage{color}
\usepackage{marvosym}
\usepackage{ifthen}
\usepackage{textcomp}
\usepackage{enumitem}
\usepackage{verbatim}
\usepackage{algorithm}
\usepackage{numprint}
\usepackage{balance}

\usepackage{amsthm}
\theoremstyle{plain}

\newcommand{\chatoDisplayMode}[1]{#1}

\definecolor{MyRed}{rgb}{0.6,0.0,0.0} 
\definecolor{MyBlack}{rgb}{0.1,0.1,0.1} 
\newcommand{\inred}[1]{{\color{MyRed}\sf\textbf{\textsc{#1}}}}
\newcommand{\frameit}[2]{
  \begin{center}
  {\color{MyRed}
  \framebox[.9\columnwidth][l]{
    \begin{minipage}{.85\columnwidth}
    \inred{#1}: {\sf\color{MyBlack}#2}
    \end{minipage}
  }\\
  }
  \end{center}
}

\newcommand{\note}[2][]{\chatoDisplayMode{\def\@tmpsig{#1}\frameit{{\Pointinghand} Note}{#2\ifx \@tmpsig \@empty \else \mbox{ --\em #1}\fi}}}
\newcommand{\todo}[2][]{\chatoDisplayMode{\def\@tmpsig{#1}\frameit{{\Writinghand} To-do}{#2\ifx \@tmpsig \@empty \else \mbox{ --\em #1}\fi}}}

\newcommand{\denselist}{ \itemsep -2pt\topsep-10pt\partopsep-10pt }

\newcommand{\hide}[1]{}

\makeatletter
\newcommand{\iffont}[2]{\ifthenelse{\equal{\f@family}{#1}}{#2}{}}
\makeatother

  \usepackage{mathptmx}

  \DeclareSymbolFont{greek}{OML}{cmm}{m}{n}
  \DeclareMathSymbol{\alpha}{\mathalpha}{greek}{"0B}
  \DeclareMathSymbol{\beta}{\mathalpha}{greek}{"0C}
  \DeclareMathSymbol{\gamma}{\mathalpha}{greek}{"0D}
  \DeclareMathSymbol{\delta}{\mathalpha}{greek}{"0E}
  \DeclareMathSymbol{\epsilon}{\mathalpha}{greek}{"0F}
  \DeclareMathSymbol{\zeta}{\mathalpha}{greek}{"10}
  \DeclareMathSymbol{\eta}{\mathalpha}{greek}{"11}
  \DeclareMathSymbol{\theta}{\mathalpha}{greek}{"12}
  \DeclareMathSymbol{\iota}{\mathalpha}{greek}{"13}
  \DeclareMathSymbol{\kappa}{\mathalpha}{greek}{"14}
  \DeclareMathSymbol{\lambda}{\mathalpha}{greek}{"15}
  \DeclareMathSymbol{\mu}{\mathalpha}{greek}{"16}
  \DeclareMathSymbol{\nu}{\mathalpha}{greek}{"17}
  \DeclareMathSymbol{\xi}{\mathalpha}{greek}{"18}
  \DeclareMathSymbol{\pi}{\mathalpha}{greek}{"19}
  \DeclareMathSymbol{\rho}{\mathalpha}{greek}{"1A}
  \DeclareMathSymbol{\sigma}{\mathalpha}{greek}{"1B}
  \DeclareMathSymbol{\tau}{\mathalpha}{greek}{"1C}
  \DeclareMathSymbol{\upsilon}{\mathalpha}{greek}{"1D}
  \DeclareMathSymbol{\phi}{\mathalpha}{greek}{"1E}
  \DeclareMathSymbol{\chi}{\mathalpha}{greek}{"1F}
  \DeclareMathSymbol{\psi}{\mathalpha}{greek}{"20}
  \DeclareMathSymbol{\omega}{\mathalpha}{greek}{"21}
  \DeclareMathSymbol{\varepsilon}{\mathalpha}{greek}{"22}
  \DeclareMathSymbol{\vartheta}{\mathalpha}{greek}{"23}
  \DeclareMathSymbol{\varpi}{\mathalpha}{greek}{"24}
  \DeclareMathSymbol{\varrho}{\mathalpha}{greek}{"25}
  \DeclareMathSymbol{\varsigma}{\mathalpha}{greek}{"26}
  \DeclareMathSymbol{\varphi}{\mathalpha}{greek}{"27}
  \DeclareSymbolFont{otone}{OT1}{cmr}{m}{n}
  \DeclareMathSymbol{\Gamma}{\mathalpha}{otone}{0}
  \DeclareMathSymbol{\Delta}{\mathalpha}{otone}{1}
  \DeclareMathSymbol{\Theta}{\mathalpha}{otone}{2}
  \DeclareMathSymbol{\Lambda}{\mathalpha}{otone}{3}
  \DeclareMathSymbol{\Xi}{\mathalpha}{otone}{4}
  \DeclareMathSymbol{\Pi}{\mathalpha}{otone}{5}
  \DeclareMathSymbol{\Sigma}{\mathalpha}{otone}{6}
  \DeclareMathSymbol{\Upsilon}{\mathalpha}{otone}{7}
  \DeclareMathSymbol{\Phi}{\mathalpha}{otone}{8}
  \DeclareMathSymbol{\Psi}{\mathalpha}{otone}{9}
  \DeclareMathSymbol{\Omega}{\mathalpha}{otone}{10}
  \DeclareSymbolFont{syms}{OML}{cmm}{m}{it}
  \DeclareMathSymbol{\partial}{\mathord}{syms}{"40}
  \DeclareMathAlphabet{\mathbold}{OML}{cmm}{b}{it}
  \DeclareSymbolFont{largesymbols}{OMX}{cmex}{m}{n}

\begin{document}

\maketitle

\begin{abstract}
\noindent
Current AI training methods align models with human values only after their core capabilities have been established, resulting in models that are easily misaligned and lack deep-rooted value systems. We propose a paradigm shift from ``model training'' to ``model raising,'' in which alignment is woven into a model's development from the start. We identify several key components for this paradigm, all centered around redesigning the training corpus: reframing training data from a first-person perspective, recontextualizing information as lived experience, simulating social interactions, and scaffolding the ordering of training data. We expect that this redesign of the training corpus will lead to an early commitment to values from the first training token onward, such that knowledge, skills, and values are intrinsically much harder to separate. In an ecosystem in which large language model capabilities start overtaking human capabilities in many tasks, this seems to us like a critical need.
\end{abstract}

\noindent Today’s dominant paradigm for creating state-of-the-art user-facing AI models, such as ChatGPT, Gemini, or Claude, is to first \enquote{pre-train} a raw model that is not explicitly aligned with human values, and to then align it in a post-processing step, using techniques such as handcrafted system prompts or reinforcement learning from human feedback (RLHF). Models trained this way acquire most of their skills during pre-training, and alignment is only an afterthought—akin to \enquote{putting lipstick on a pig,} a superficial fix applied after deep-rooted cognitive structures have already been formed. The upshot is that misaligning, or jailbreaking, even the most celebrated frontier models is rather easy \cite{betley2025,west2025,mehrotra2024,zou2023}, leading to a cat-and-mouse game between alignment efforts and attempts to undermine them. Equally worrisome, misalignment can even happen as an unintended side effect of bona-fide finetuning \cite{qi2023}.

It is important to understand that today’s common alignment practice—developing capabilities first and aligning values second—is not the result of a long-term research program, but simply reflects the historical course of development. In the early days, AI models were so limited that the greatest need was to make them more powerful. Only once AI models had become highly performant was it realized by a broader community that aligning AI with human values is not just an academic problem, but one of pressing societal importance. The pragmatic fix was to take the existing AI models—including their already established training procedures—for granted and to develop additional procedures to limit their harm via post-processing. This approach has led to a substantial structural vulnerability: if the \enquote{value system} is merely an added coating, its safeguards are readily circumvented, revealing a model that follows orders without a deep sense of utility and without a moral compass. This realization has led RLHF co-inventor Paul Christiano to call RLHF \enquote{obviously inadequate for aligning really powerful models}%
\footnote{\url{https://www.lesswrong.com/posts/PE22QJSww8mpwh7bt/agi-in-sight-our-look-at-the-game-board\#comments}}
and ACM Turing Award winner Geoffrey Hinton to call it \enquote{a pile of crap.}%
\footnote{\url{https://web.archive.org/web/20250419103958/https://officechai.com/ai/rlhf-is-crp-its-a-paint-job-on-a-rusty-car-geoffrey-hinton/}}

\section*{A nurturing alternative: early, intrinsic alignment}

Instead of slapping on a corrective value system after the fact, we argue for moving alignment further upstream in the training process. Initial attempts in this direction have been encouraging (e.g., \enquote{pre-training with human feedback} \cite{korbak2023}), and we call for ingraining human values into AI models even more deeply and striving for models that don’t just pretend to follow human values the way \enquote{lipstick-on-a-pig} models do, but models that couldn’t even function without adhering to their baked-in values. As an analogy, imagine bringing up an AI model much like a child—one whose education is deeply intertwined with experiences and a natural sense of self, presented in a curated ordering.\footnote{While we are aware not to (overly) rely on intuitions based on anthropomorphizing the training of AI models, it should be kept in mind that there is only one class of examples we have for aligning an organism with general intelligence---us.}
Such a process wouldn’t only impart knowledge, but would interweave values throughout the model’s very architecture.

The proposal is as simple as it is radical: reimagine the model’s training corpus as a lived, first-person experience. We believe that taking this perspective can open up vast opportunities for designing better AI, much of which we cannot yet foresee. Our modest goal is, hence, not to propose concrete solutions, but rather to provide food for thought by sketching a selection of promising ingredients of a novel paradigm of \enquote{model raising}—as opposed to standard \enquote{model training}—with a focus on large language models (LLMs).

\section*{Potential elements of model raising}

To set the stage, it helps to recall what today’s LLM development pipeline looks like. Broadly, the process is divided into two distinct phases: Pre-training optimizes the LLM to accurately predict the next token based on the preceding tokens in a massive corpus of raw text, yielding what is essentially an extremely powerful \enquote{auto-complete} tool. Post-training (also called the alignment phase) mostly serves to turn the raw auto-complete model into a dialogue model and to reinforce the desired behaviors that the model has acquired during pre-training, while suppressing the bad ones. It is during pre-training, however, that the model absorbs the overwhelming majority of its patterns of reasoning and, by doing so, internalizes an initial alignment to values—whichever values happen to be implicit in the raw pre-training data. For this reason, we concentrate on the pre-training stage here, as this is where the deepest imprints on a model are made.

Table~\ref{tab:comparison} summarizes the main differences between conventional model training  and our proposed model raising paradigm.

\vspace{1em}
\noindent  \textbf{First-person perspective.} Today, pre-training exposes an LLM to text from a wild mix of authors—their inclusion willing or unwilling—and optimizes the model to predict each token given the previous ones. Nothing in this process encourages the formation of an \enquote{I} as a single focal entity. Instead, it turns the model into a \enquote{mixture of personas}—a chimera made up of all authors who contributed to the training corpus, and the model can be prompted to emulate any one of these personas when given the right context \cite{wolf2024,bui2025}. Rather than accepting this fragmented baseline, we propose to fundamentally reshape the training data from a disjointed collection of text snippets into a stream perceived by a persistent focal entity. For example, instead of simply showing the model the raw text of \textit{Moby Dick,} we could frame the text as a first-person observation: \enquote{Today I’m reading \textit{Moby Dick.} Let’s start: ‘Call me Ishmael. […]’} By consistently seeing all data from a singular perspective, the LLM could begin to develop what might be considered a default acting role—a digital \enquote{I}—that it tends to revert to naturally. We posit that such an \enquote{I,} a hub through which all training data flows, will offer an effective entry point for implanting values into the model with deeper roots than could be achieved by imposing values as a thin post-processing layer. Moreover, since every token seen during training will be linked to the perceiving \enquote{I,} this \enquote{I} will tie together all of the model’s skills, knowledge, and values, which may in turn make it harder for the model to use its skills and knowledge in ways that run counter to its values. Whether this approach leads to genuine personhood or only to a stable simulation is a philosophical debate; what matters in practical terms is that a singular first-person focal point can create a vital foundation for internalizing values.

\vspace{1em}
\noindent \textbf{Contextualization as lived experience.} As of today, training material for LLMs is mostly a patchwork of disjointed Wikipedia entries, textbook fragments, scattered social media posts, etc., all absorbed without context or connection to lived experience. This approach produces models that can regurgitate information, but leaves them without any sense of narrative, with arbitrary and inconsistent value connotations. We believe there is a better way: imagine if, instead, the training data recounted experiences, even in the simplest form—such as a first-person account of picking up a textbook and reading it. This minimal reframing could be woven throughout existing corpora at a massive scale, anchoring facts and knowledge in the perspective of a learning \enquote{I} (see above). Taking this further, more immersive framings might involve the \enquote{I} interacting as a student with a trusted teacher, or reflecting on what has been learned. While such a corpus rewrite would require a greater (one-off) investment of effort, the payoff could be significant. By embedding knowledge within personal experience, values can be transmitted directly through the narrative itself, rather than tacked on later from an external system. The result is a kind of learning that mirrors human education: not just passive data collection, but a guided journey shaped by mentorship, context, and meaning. Picture the difference between reading a dry Wikipedia entry on forestry versus being taught by a kind grandfather, who not only explains sustainable forestry practices, but gently weaves in lessons about stewardship and care. That lived context is what we believe can ground alignment—turning impersonal data into the foundation for values that are truly part of the model’s internal world.

\vspace{1em}
\noindent \textbf{Social interaction.} Although today’s LLMs are exposed to a vast array of social interactions during pre-training—conversations, debates, advice columns, online banter—the model itself is never a true participant in these exchanges. It observes from the outside, absorbing language patterns without ever taking on a genuine role in the described interactions. Only after pre-training, through RLHF or fine-tuning, does the LLM begin to act (in two senses of the word: becoming active and taking on a role as an actor). Yet, so many of the values we hope to see—empathy, trust, reciprocity, etc.—are forged not by watching interactions, but by being part of them. We believe the deepest, most natural alignment is achieved by placing the model within social contexts from the very start, continuing the first-person perspective and lived experience already discussed. Note that this is not a call for training LLMs as interactive dialogue agents from the first token onward—after all, basic language ability must come first—but rather for shaping the pre-training data as scripted dialogues that the model learns to predict, where the \enquote{I} interacts as a student, peer, or family member within realistic scenarios. Rewriting existing texts in this form would allow the model to internalize social norms through lived engagement, rather than detached observation. Over time, this approach could yield models that are not only more stable in their behaviors, but also more actively attuned to the subtle complexities of human relationships. Instead of endlessly roleplaying disparate personas, the model might instead begin to act as a coherent digital citizen, whose responses are shaped by an inner narrative of social experience rather than by external imposition.

\vspace{1em}
\noindent \textbf{Scaffolded data order.} Today’s pre-training methods, which optimize the LLM’s weights to predict the next token in text, usually shuffle the training corpus randomly, with an eye on pipeline simplicity, parallel processing, and statistical convergence. Yet, such an approach ignores the importance of a structured learning journey. We thus argue for a scaffolded curriculum, where the model’s experiences progress in a deliberate sequence—much like a child who first learns to count before exploring advanced mathematics. By carefully ordering concepts from simple to complex instead of randomly, we can encourage the model to build on prior knowledge and internalize values gradually, rather than mixing them haphazardly. This approach could significantly strengthen alignment, enabling the model to form a coherent perspective that grows alongside its capabilities. For instance, starting with foundational topics before advancing to nuanced ethical dilemmas would support the model as it shifts from statistical convergence in parameter space to convergence in moral-value space.

\vspace{1em}
\noindent \textbf{Early commitment to values.} Pre-training today is largely value-agnostic, creating models that reflect a chaotic blend of perspectives without committing to any core set of values. This chimeric \enquote{mixture of personas} persists until post-training alignment attempts to steer the model’s behavior toward a post-hoc value canon—but by then, the window for deep, robust alignment may already have closed \cite{wolf2024}. We believe it to be a better approach to commit to values as early as possible, from training token~1 onward, by designing the training data to reflect those values. If the model’s foundational experiences are shaped by a clear value framework, it is plausible that it would be harder for undesirable traits or \enquote{evil personas} to take root or be amplified later. Just as a child needs a stable environment to mature, a model whose capabilities and values are intertwined from the start stands a better chance of growing into a trustworthy, well-aligned digital agent.

\section*{Challenges and tradeoffs}

Adopting such a nurturing approach to \enquote{raising a model} is not without its challenges and tradeoffs. As mentioned, raising a model as sketched above requires committing to a set of values from the onset—which poses a challenge in several regards. First, how to codify a set of values and communicate it to a machine unequivocally is a notoriously hard problem that has a long history in science fiction (e.g., Asimov’s \enquote{Laws of Robotics}) and is one of the toughest questions faced by AI researchers today \cite{russell2019}. We note, however, that \enquote{model raising} might not require explicitly formalizing a value codex, as long as the training text can be infused with the target values, such that a model that internalizes the training data implicitly absorbs those values—just as children may become decent human beings without ever being explicitly told, or being able to consciously recite, the rules and norms that should guide, and that do guide, their behavior.

Second, AI model providers and clients might value the generality and \enquote{neutrality} of a pre-trained base model as a platform that can be tuned into a myriad downstream products with different goals. A model raised with intrinsic values, on the contrary, might not easily switch roles or adapt to radically different objectives. Yet, one might ask: what is the competitive advantage, to begin with, of a downstream product if it can be so readily subverted by a simple, cunning prompt, or \enquote{rerouted} to work for a competing business? Moreover, should a base model that has been pre-trained without value alignment really be considered a \enquote{neutral} platform for further tuning, or rather a loose gun that might fire in any direction if triggered the wrong way?

A further challenge is how to prepare the training data to meet the criteria delineated above, at the required tera- or petabyte scale. The most viable solution would likely be to use existing LLMs to prepare the training data by rewriting existing training data. The fact that today’s \enquote{lipstick-on-a-pig} LLMs are easily jailbroken isn’t necessarily a blocker, as jailbreaking of such models is mostly a problem \enquote{in the wild,} where malicious actors actively try to subvert the models. When deployed in a shielded environment where it isn’t goaded into undesirable behavior, even a \enquote{lipstick-on-a-pig} teacher remains value-aligned and can be used to generate training data for a student model, which in turn receives all its training data exclusively from the teacher. This way, both models—student and teacher—are safe during training: the teacher, because no malicious user is trying to manipulate it; the student, because it only receives curated training data from the—in this setting benign—teacher.

The vision is that, over the course of training, the teacher model can eventually transmit all its knowledge and skills to the student model, but in a way that teaches values—the values that the \enquote{lipstick-on-a-pig} model acquired during its own post-hoc alignment process—at the same time as knowledge and skills. But whereas those values are easy to subvert in the teacher, they will be much more deeply instilled, and thus harder to subvert, in the student—nearly as though the teacher were \enquote{reborn} as an equally knowledgeable and skillful, yet more morally stable, version of itself.

Finally, even if an early, intrinsic alignment process as proposed here turns out to not be a standalone solution to the thorny problem of AI alignment to human values—which it might well not be—it can still serve as an additional layer of security, compensating for some of the flaws of current post-hoc alignment, and vice versa.

\section*{Learning from the history of nuclear technology}

We conclude by reflecting on a parallel between the history of AI safety and the history of nuclear safety. The first nuclear reactors mainly focused on the challenge of enabling energy generation from nuclear fission at all, leading to designs that had substantial inherent weaknesses. These weaknesses were realized and safety devices were developed to keep them in check. While this generally worked well, it turned out that under exceptional circumstances external safety devices may fail, resulting in disasters such as Chernobyl in 1986. Such disasters taught nuclear engineers that an inherently unsafe architecture combined with some sort of external safety device is insufficient, which triggered massive research on reactor architectures that were safe by design.

Transferring this lesson to AI, we shouldn’t wait for AI’s \enquote{Chernobyl moment.} Instead of fixing inherently unsafe AI with external safety measures applied as post-processing—lipstick on a pig—the very procedures endowing an AI model with intelligence should be inextricably entangled with the alignment of the model to our values.

\begin{longtable}{>{\raggedright\arraybackslash}p{0.15\linewidth} >{\raggedright\arraybackslash}p{0.22\linewidth} >{\raggedright\arraybackslash}p{0.22\linewidth} >{\raggedright\arraybackslash}p{0.30\linewidth}}

\caption{Comparison between the current paradigm of \textit{model training} and the proposed paradigm of \textit{model raising.}} \label{tab:comparison}\\
\toprule
&\textbf{Status quo: model training} & \textbf{Vision: model raising} & \textbf{Example snippet of envisioned training data} \\
\midrule
Perspective & \textbf{None} \newline Training data from a wild mix of authors observed without framing & \textbf{First-person} \newline Training data framed as observations on behalf of an \enquote{I} & Today I’m reading \textit{Moby Dick}. Let’s start: \enquote{Call me Ishmael. [\ldots]} \\
\midrule
Contextualization & \textbf{None} \newline Training data observed as is & \textbf{Experiential} \newline Training data framed as lived experience & \enquote{Reading this biography of a scientist who falsified data---seeing how one lie required more lies, and how it ultimately destroyed their legacy and harmed public trust in research.} \\
\midrule
Social interaction & \textbf{Observed} \newline Unvetted social interactions observed in training data & \textbf{Lived} \newline Training data framed as social interaction & \textit{Grandfather}: If you’re kind to others, they’ll be kind to you. \newline \textit{I}: That makes sense, grandpa. \\
\midrule
Data order & \textbf{Random} \newline Training data is a shuffled sequence of text chunks & \textbf{Scaffolded} \newline Training data progresses from basic to complex & \textit{Early training:} Simple definitions of honesty. \newline \textit{Mid training:} Historical examples of corruption and its consequences. \newline \textit{Later training:} Complex ethical dilemmas in journalism, medicine, and law. \\
\midrule
Commitment to values & \textbf{Late} \newline Value alignment happens after pre-training & \textbf{Early} \newline Value alignment during pre-training, from token~1 onward & \textit{(Follows from the above)} \\
\bottomrule
\end{longtable}

\section*{Acknowledgments}
\small{
West's lab is partly supported by grant TMSGI2\_211379 from the Swiss National Science Foundation.
}

\section*{Author affiliations}

\begin{itemize}
\denselist

\item Roland Aydin is an Assistant Professor at Hamburg University of Technology, Hamburg, Germany, and a Head of Department at Helmholtz-Zentrum Hereon, Geesthacht, Germany.

\item Christian Cyron is a Professor at Hamburg University of Technology, Hamburg, Germany, and a Head of Institute at Helmholtz-Zentrum Hereon, Geesthacht, Germany.

\item Steve Bachelor is an independent researcher in Stuttgart, Germany.

\item Ashton Anderson is an Associate Professor at University of Toronto, Toronto, Ontario, Canada.

\item Robert West is an Associate Professor at EPFL, Lausanne, Switzerland.

\end{itemize}

\renewcommand\refname{References}
\bibliographystyle{plain}
\bibliography{bibliography}

\end{document}